\documentclass[conference,letterpaper,twoside,final,10pt]{ieeeconf}

\usepackage{graphicx}
\usepackage{amssymb}
\usepackage{amsmath}
\usepackage{cite}

\usepackage{color}
\usepackage{url}
\usepackage{alltt}
\usepackage{cleveref}
\usepackage{tikz}
\usepackage{pgfplots}
\usepackage{pgf-umlsd}
\usepackage{cprotect}
\usepackage{multirow}
\usepackage{algorithm}
\usepackage[noend]{algpseudocode}
\usepackage{bigfoot}
\usepackage{paralist}
\usepackage{tabularx}
\usepackage[whole]{bxcjkjatype}
\usepackage{ifthen}
\usepackage{threeparttable}
\usepackage[hang,small,bf]{caption}
\usepackage[subrefformat=parens]{subcaption}
\pgfplotsset{compat=1.14}
\IEEEoverridecommandlockouts
\DeclareMathOperator*{\argmin}{argmin}

\title{\LARGE \bf
Interactively Picking Real-World Objects with\\
Unconstrained Spoken Language Instructions
}

\author{
Jun Hatori$^{*}$,
Yuta Kikuchi$^{*}$,
Sosuke Kobayashi$^{*}$,
Kuniyuki Takahashi$^{*}$,\\
Yuta Tsuboi$^{*}$,
Yuya Unno$^{*}$,
Wilson Ko,
Jethro Tan$^{\dagger}$
\thanks{$^{*}$ The starred authors are contributed equally and ordered alphabetically.}
\thanks{$^{\dagger}$ All authors are associated with Preferred Networks, Inc.
{\{hatori, kikuchi, sosk, takahashi, tsuboi, unno, wko, jettan\}@preferred.jp}}}

\begin{document}

\maketitle
\thispagestyle{empty}

\begin{abstract}%

Comprehension of spoken natural language is an essential skill for robots to communicate with humans effectively.
However, handling unconstrained spoken instructions is challenging due to (1) complex structures and the wide variety of expressions used in spoken language, and (2) inherent ambiguity of human instructions.
In this paper, we propose the first comprehensive system for controlling robots with unconstrained spoken language, which is able to effectively resolve ambiguity in spoken instructions.
Specifically, we integrate deep learning-based object detection together with natural language processing technologies to handle unconstrained spoken instructions, and propose a method for robots to resolve instruction ambiguity through dialogue.
Through our experiments on both a simulated environment as well as a physical industrial robot arm, we demonstrate the ability of our system to understand natural instructions from human operators effectively, and show how higher success rates of the object picking task can be achieved through an interactive clarification process.
\footnote{Accompanying videos are available at the following links:\\ \url{https://youtu.be/_Uyv1XIUqhk} (the system submitted to ICRA-2018) and \url{http://youtu.be/DGJazkyw0Ws} (with improvements after ICRA-2018 submission)}
\end{abstract}

\section{Introduction}
\label{sec:introduction}
As robots become more omnipresent, there is also an increasing need for humans to interact with robots in a handy and intuitive way.
For many real-world tasks, use of spoken language instructions is more intuitive than programming, and is more versatile than alternative communication methods such as touch panel user interfaces~\cite{ochiai2014remote} or gestures~\cite{shukla2015probabilistic} due to the possibility of referring to abstract concepts or the use of high-level instructions.
Hence, using natural language as a means to interact between humans and robots is desirable.

\begin{figure}
\vspace{5mm}
\centering
\includegraphics[width=\columnwidth]{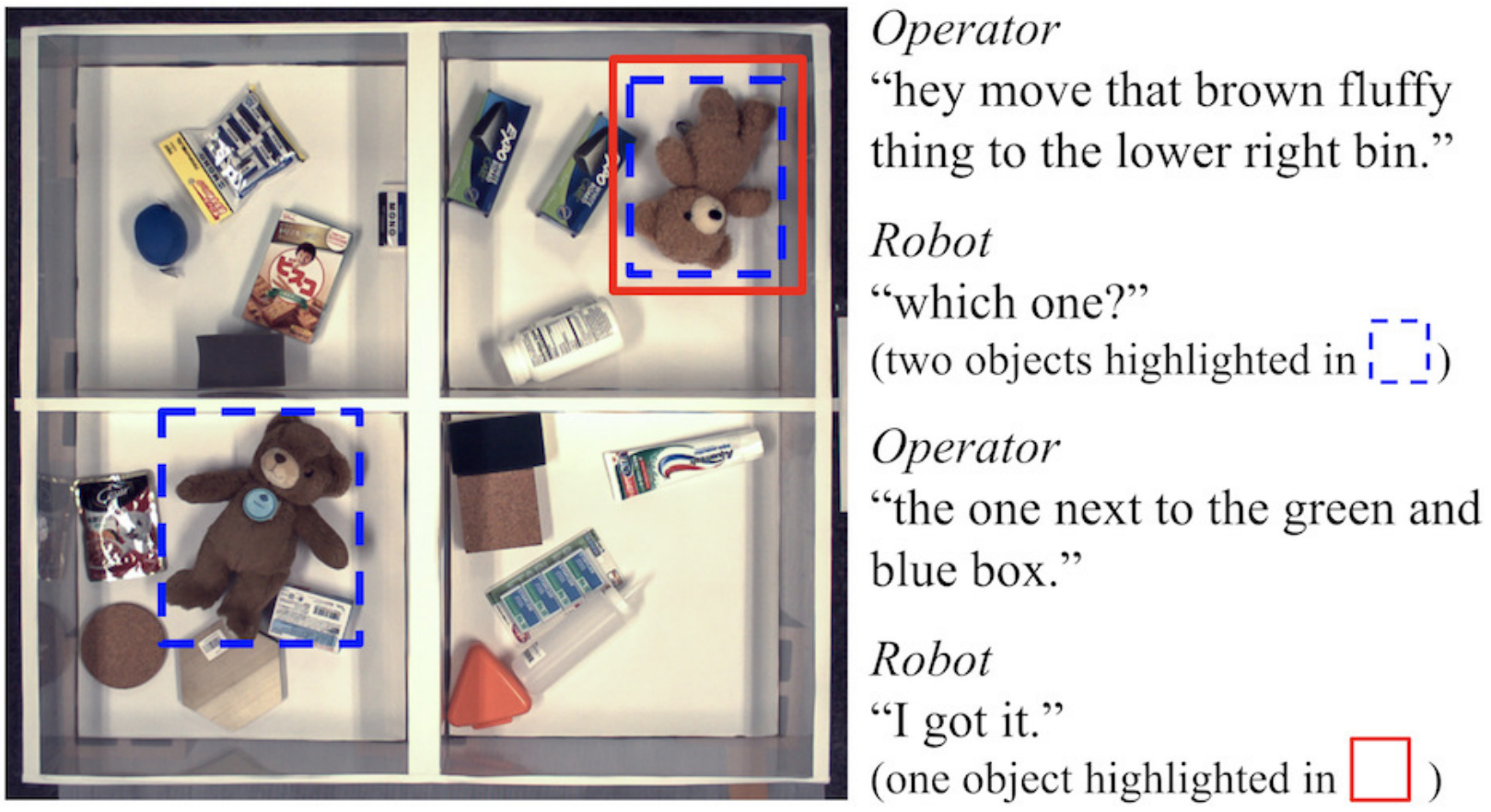}
\caption{An illustration of object picking via human--robot interaction. Our robot asks for clarification if the given instruction has interpretation ambiguity.}
\label{fig:interaction}
\end{figure}

However, there are two major challenges to realize the concept of \emph{robots that interpret language and act accordingly}.
First, spoken language instructions as used in our daily lives have neither predefined structures nor a limited vocabulary, and often include uncommon and informal expressions, e.g., ``Hey man, grab that brown fluffy thing'', see~\Cref{fig:interaction}.
Second, there is inherent ambiguity in interpreting spoken languages, since humans do not always put effort in making their instructions clear.
For example, there might be multiple ``fluffy'' objects present in the environment like in~\Cref{fig:interaction}, in which case the robot would need to ask back: e.g. ``Which one?''.
Although proper handling of such diverse and ambiguous expressions is a critical factor towards building domestic or service robots, little effort has been made to date to address these challenges, especially in the context of human--robot interaction.

In this paper, we tackle these two challenges in spoken human--robot communication, and develop a robotic system which a human operator can communicate with using unconstrained spoken language instructions.
To handle complex structures and cope with the diversity of unconstrained language, we combine and modify existing state-of-the-art models for object detection~\cite{liu2016ssd,alexe2012obj} and object-referring expressions~\cite{yu2016modeling,yu2017joint} into an integrated system that can handle a wide variety of spoken expressions and their mapping to miscellaneous objects in a real-world environment.
This modification makes it possible to train the network without explicit object class information, and to realize zero-shot recognition of unseen objects.
To handle inherent ambiguity in spoken instructions, our system also focuses on the process of interactive clarification, where ambiguity in a given instruction can be resolved through dialogue.
Moreover, our system agent combines verbal and visual feedback as shown in~\Cref{fig:interaction} in such a way that the human operator can provide additional explanations to narrow down the object of interest similar to how humans communicate.
We show that spoken language instructions are indeed effective in improving the end-to-end accuracy of real-world object picking.

Although the use of natural language instructions has received attention in the field of robotics~\cite{paul2016efficient, shridhar2017grounding, whitney2017reducing,dang2016interpreting}, our work is the first to propose a comprehensive system integrating the process of interactive clarification while supporting unconstrained spoken instructions through human--robot dialogue.
To evaluate our system in a complex, realistic environment, we have created a new challenging dataset which includes a wide variety of miscellaneous real-world objects, each of which is annotated with spoken language instructions.

The remainder of this paper is organized as follows.
Related work is described in~\Cref{sec:relatedwork}, while~\Cref{sec:Task Definition} explains our task definition and~\Cref{sec:method} explains our proposed method.
\Cref{sec:experiment} outlines our experimental setup, while results are presented in~\Cref{sec:results}.
Finally, \Cref{sec:conclusion} describes our future work and concludes this paper.

\section{Related Work}
\label{sec:relatedwork}
\subsection{Human--Robot Interaction with Natural Language}
Natural language processing has recently been receiving much attention in the field of robotics.
Several studies on human--robot interactions have been conducted, such as those focusing on the expressive space of abstract spatial concepts as well as notions of cardinality and ordinality~\cite{paul2016efficient, shridhar2017grounding}, or those focusing on how to ask a robot questions to clarify ambiguity~\cite{whitney2017reducing}.
The work in~\cite{paul2016efficient, shridhar2017grounding} shows that the positional relationship in space is essential to deal with multiple objects.
Most notably, in~\cite{shridhar2017grounding} multiple daily objects are recognized, instead of similarly shaped simple objects with different colors, as in~\cite{paul2016efficient}.
However, these studies implicitly assume that human instructions are precise, that is, without interpretation ambiguity.
As mentioned in the introduction though, natural language is inherently ambiguous.

The language-based feedback to robots, especially under the presence of interpretation ambiguity, has been overlooked except by few research work such as \cite{whitney2017reducing}.
\cite{whitney2017reducing} focused on object fetching using natural language instructions and pointing by gesture.
However, experiments were only conducted in a simple, controlled environment containing at most six known and labeled objects. 
Moreover, feedback from the robot is limited to only a simple binary confirmation: i.e. ``This one?''.
Although it is possible to execute specific tasks such as cooking~\cite{dang2016interpreting} by designing the environment and labels and teaching the robot in advance, it is difficult to respond under circumstances where a large number of miscellaneous objects are placed in the environment, especially under the presence of objects that do not have a specific or common name to call.

\subsection{Comprehension of Visual Information with Natural Language}

Recently, there has been significant progress in multi-modal research related to computer vision and natural language processing.
The capability of comprehending images has been improved and measured through image captioning~\cite{kiros2014multinlm, vinyals2015show} using neural network based models.
Mapping language information into a visual semantic space~\cite{frome2013devise} and image retrieval~\cite{kiros2014unifying} have also been investigated.
For more precise comprehension~\cite{krishna2017genome}, captioning images densely for each object~\cite{johnson2016densecap} or relation~\cite{lu2016visual} has been done.
Further work includes work on tasks for referring expressions~\cite{kazemzadeh14refer, mao2016gencomp, nagaraja2016modeling, yu2017joint}, and generating or comprehending discriminative sentences in a scene where similar objects exist.

Visual question answering~\cite{antol2015vqa} is also a related task that is answering single open-ended questions about an image.
Object detection~\cite{ren15faster,liu2016ssd}, which plays an important role in our work, has been improved through several grand challenges such as PASCAL VOC~\cite{everingham10pascalvoc} and MSCOCO~\cite{lin2014mscoco}.
In our task setting, the system tries to identify particular objects within an image described with a spoken expression.
However, it is naive to assume that the referred object is always uniquely defined in a natural conversation between human and robots.
Therefore, we introduce an interactive system to clarify target objects using spoken language.

\section{Task Definition}
\label{sec:Task Definition}
As described in~\Cref{sec:introduction}, our work focuses on the interaction between a human operator and a system agent controlling a robotic system through natural language speech input.
To this extent, we set up an environment in which over 100 objects are distributed across several boxes as shown in~\Cref{fig:environment}.
The robot is then instructed by the operator to pick up a specific object and move it to another location, for example in the following manner: \textit{``Move the tissue box to the top right bin.''}

One part of the interaction we focus on in this work is the process of getting more clarity.
Looking again at~\Cref{fig:environment}, for example, two tissue boxes are present in the workspace of the robot, which might lead the system agent to ask the operator: \textit{``Which one do I need to pick?''}.
The operator is then allowed to use unconstrained language expressions as feedback to direct the system agent to the requested item, which in our example could be: \textit{``The orange one, please.''}, \textit{``Get me the one that is in the corner.''}, or \textit{``The one next to the plastic bottle.''}.

We focus on highly realistic and challenging environments in the following aspects:
\begin{itemize}
\item Highly cluttered environment: We focus on both organized and highly cluttered environments. In our setup for cluttered environments, a large number of objects are scattered over four boxes, with a large degree of occlusion. Object recognition as well as picking in this kind of environment is very challenging, especially when a large part of the object is occluded. On the other hand, highly organized environments are also challenging. In this case the human operator might need to use various, additional expressions to refer to a unique object among similar objects in the environment. These expressions could refer to the absolute or relative position to another object, or could be ordinality/cardinality expressions.
\item Wide variety of objects: we use over a hundred different known objects, including those that do not have specific names and hence would force a human operator to use indirect or abstract expressions to refer to the target object. In addition, we used 22 unknown objects to evaluate how well our recognition and picking model generalizes to new objects.
\end{itemize}
\section{Proposed Method}
\label{sec:method}
Taking advantage of recent advances in image and referring expression comprehension models, we built a fully statistical language understanding system, which does not rely on any hand-crafted rules or pre-defined sets of vocabulary or grammatical structures.
By jointly training the object recognition and language understanding modules, our model can learn not only the general mapping between object names and actual objects, but also various expressions referring to attributes of each object, such as color, texture, size, or orientation.

\begin{figure}
\begin{center}
\includegraphics[width=\columnwidth]{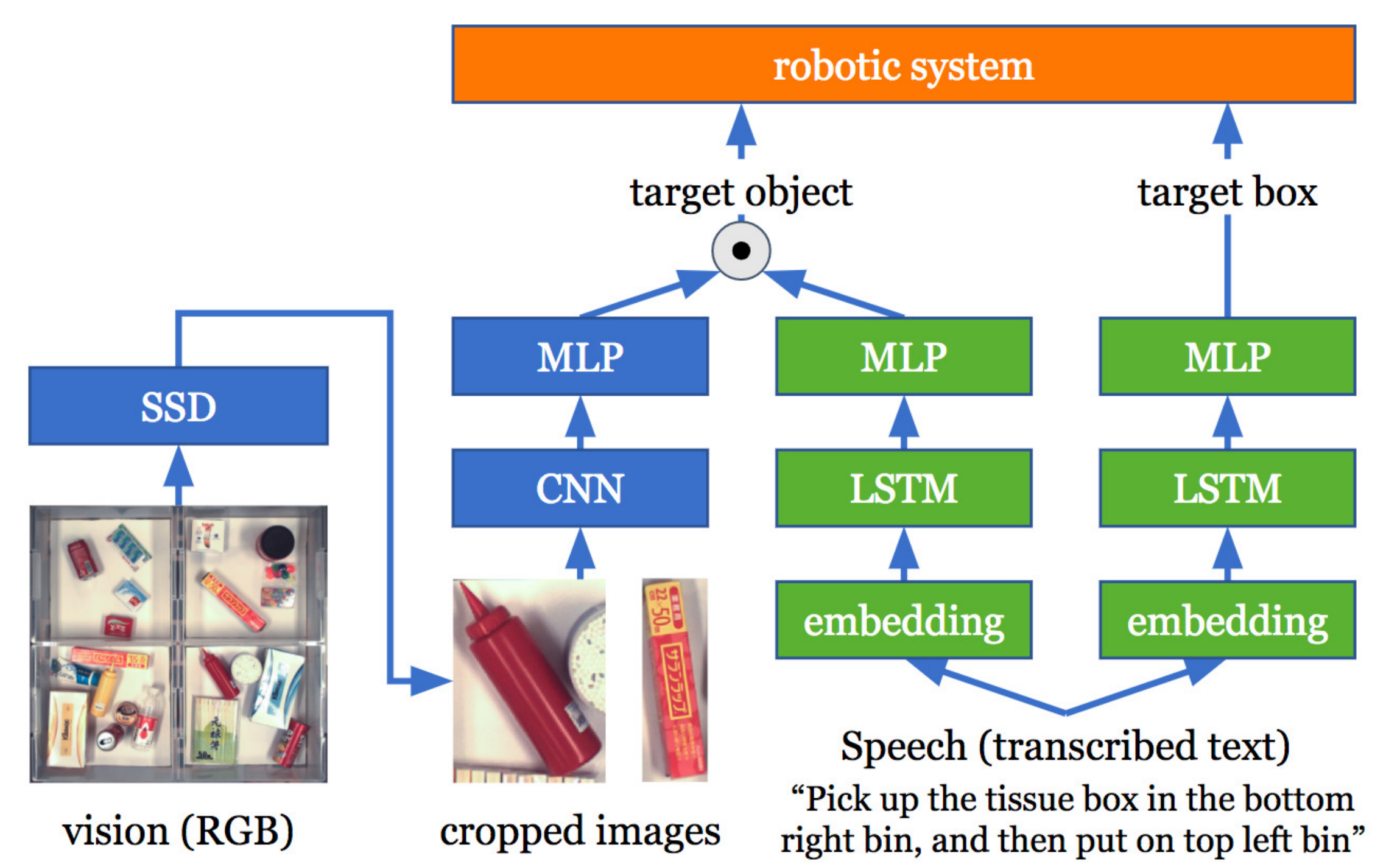}
\caption{Overview of the neural network architecture. The image input from our vision system is split into copped images using SSD, which are then fed into CNN to extract image features. Image and text features extracted from the vision and text (speech) input are fed into the target object detection module, while only the text feature is fed into the destination box selection model.}
\label{fig:network-architecture}
\end{center}
\end{figure}

Instructions from a human operator are given as text or speech input.
The goal of comprehension of the instruction (i.e. specifying the target object and the destination) is divided into five subtasks (Figure~\ref{fig:network-architecture}):
\begin{enumerate}
\item Transcribe the speech input as a textual instruction using the Web Speech API in Google Chrome (only for speech input).
\item Detect a bounding box for each object in an image taken by the robot's camera using an SSD-based detector.
\item Identify the target object by using an encoder model that takes the textual instruction and the image (within the bounding box) as input.
\item Identify the destination box with a classifier that takes the textual instruction (but not image) as input.
\item If the given instruction turns out to be ambiguous, provide feedback to the human operator and ask for elaboration.
\end{enumerate}

Details of each subtask will be described in the following sections.

\subsection{Candidate Object Detection}

To detect a bounding box for each object, we train an object detection model based on a Single Shot Multibox Detector (SSD)~\cite{liu2016ssd}, which encodes regions in the input image efficiently and effectively.
The model scores a large number of cropped regions in the image and outputs bounding boxes for candidate objects by filtering out regions with low scores.
To handle previously-unseen objects in our challenging setting, we modified the original SSD so that it simply classifies each candidate bounding box into \textit{foreground object} or \textit{background}, following \cite{alexe2012obj, ren15faster}.
This modification enables the model to be trained without explicit object class information, since each region is scored purely based on its \textit{objectness}~\cite{alexe2012obj}, and also allows us to take advantage of existing large-scale data created for different domains or object classes (e.g., VOC PASCAL~\cite{everingham10pascalvoc}, MSCOCO~\cite{lin2014mscoco}).
Our model will also be able to detect unknown objects that do not appear in the training data by generalizing over all training objects.
Our implementation of the modified SSD is built on top of reimplementation of the original SSD algorithm\footnote{\url{https://github.com/chainer/chainercv/tree/master/examples/ssd}} with Chainer~\cite{tokui2015chainer} and Chainer-CV~\cite{niitani2017chainercv}.

\subsection{Target Object Selection}
\label{sec:object_recognition}

After bounding boxes for all objects are recognized, we need to identify where the specified target object is located.
Our object comprehension module is based on a referring expression listener model~\cite{yu2017joint}, but we have additionally made a modification to support zero-shot recognition of unseen objects.

The target object recognition is formulated as a task to find the best bounding box $\hat{b}$ from a set of predicted bounding boxes $B = \{b_i\}$, given an instruction $q$ and the entire image from a view $I$.
As shown in \Cref{fig:network-architecture}, the image is first cropped into regional images defined by the bounding boxes of the candidate objects, and each object image is encoded with a convolutional neural network (CNN), followed by a multi-layer perceptron (MLP) layer that combines geometric features of each bounding box and relational features comparing the geometric features with those of other objects (i.e., average-/max-/min-pooling over normalized differences with the objects).
On the other hand, the instruction $q$ is encoded in a word embedding layer, which encodes each word in the input sentence to a vector representation, followed by Long Short-Term Memory (LSTM) and MLP layers.
Finally, the output module calculates scores $\{s_i | s_i \in [-1, 1]\}$ for each object by calculating the cosine similarity between the vector representation of the instruction and that of the cropped object image.

When calculating the relational features, our module compares the geometric features with all other objects in the environment, rather than with those in the same object class, as was done in the original model by \cite{yu2017joint}.
This modification allows us to build a generalized model that can recognize both seen and unseen objects, and also simplifies our data creation process as we do not need to annotate each object with its object class.

\subsection{Destination Box Selection}
Our system agent also needs to identify the destination box to which it will move the target object.
As shown in \Cref{fig:network-architecture}, our destination box selection module employs the same neural network architecture as used for \Cref{sec:object_recognition}: a neural network consisting of a word embedding layer, LSTM layer, and MLP layer.
Although we separate the tasks of target object recognition and destination box selection, we found that the two tasks can actually be solved with the same architecture, and we do not have to perform tokenization or chunking of the instruction sentence.
With an LSTM-based model, each network learns which part of the input sentence should be focused on, and the two recognition models will be trained to extract different types of information (i.e., target object and destination box) even from the same instruction sentence.
We found that our LSTM-based encoder, which does not rely on any hand-crafted rules, is particularly effective for spoken language instructions, which are informal and often contain ungrammatical structures or word order.

\subsection{Handling Ambiguity}
In cases where the instruction from the human operator is unclear or erroneous, the system agent needs to ask for clarification.
Formally we define this as the process to uniquely identify the target object and destination box with high confidence.
We use a simple margin-based method: the system agent assumes it has confidently identified the target object and box if the score is higher than those of other objects (or boxes) with a margin larger than a threshold $m_{obj}$ and $m_{box}$.
If more than one object or box has scores within the threshold, we consider all objects or boxes within the margin to be potential targets, letting the system agent ask the human operator to provide additional explanation while highlighting all objects or boxes in the display, as in \Cref{fig:interaction}.

Tuning margin $m$ is hard work and critical, because too large a margin causes persistent confirmation and too small a margin causes speculative actions even when ambiguity exists.
However, we propose to exploit a reasonable margin which is linked to the training of a comprehension model.
To train our comprehension model $f_{\theta}$, which scores a pair of a sentence and an object, we minimize a max-margin loss between scores of positive and negative object--sentence pairs.
That is,
\begin{equation*}
\begin{split}
\argmin\limits_{\theta}
\mathbb{E}_{q, o} \large[
&\max \{ 0, m - f_{\theta}(q, o) + f_{\theta}(q, \hat{o}) \} + \\
&\max \{ 0, m - f_{\theta}(q, o) + f_{\theta}(\hat{q}, o) \}
\large],
\end{split}
\end{equation*}
where $q$ and $\hat{q}$ are correct and incorrect sentences, and $o$ and $\hat{o}$ are correct and incorrect objects.
In other words, this training aims to guarantee that every correct pair of a sentence and an object has scores higher by a margin $m$ than any other pair with a wrong sentence or object.

If additional sentences are fed, the model simply calculates the scores of objects by summing scores using each sentence.
We found that this approach is effective, and demonstrate so in our experiments~\Cref{sec:results}.

\section{Experiment Setup}
\label{sec:experiment}

\subsection{Dataset}

\begin{figure}
\centering
\vspace{2.5mm}
\begin{minipage}{0.47\hsize}
\includegraphics[width=\hsize]{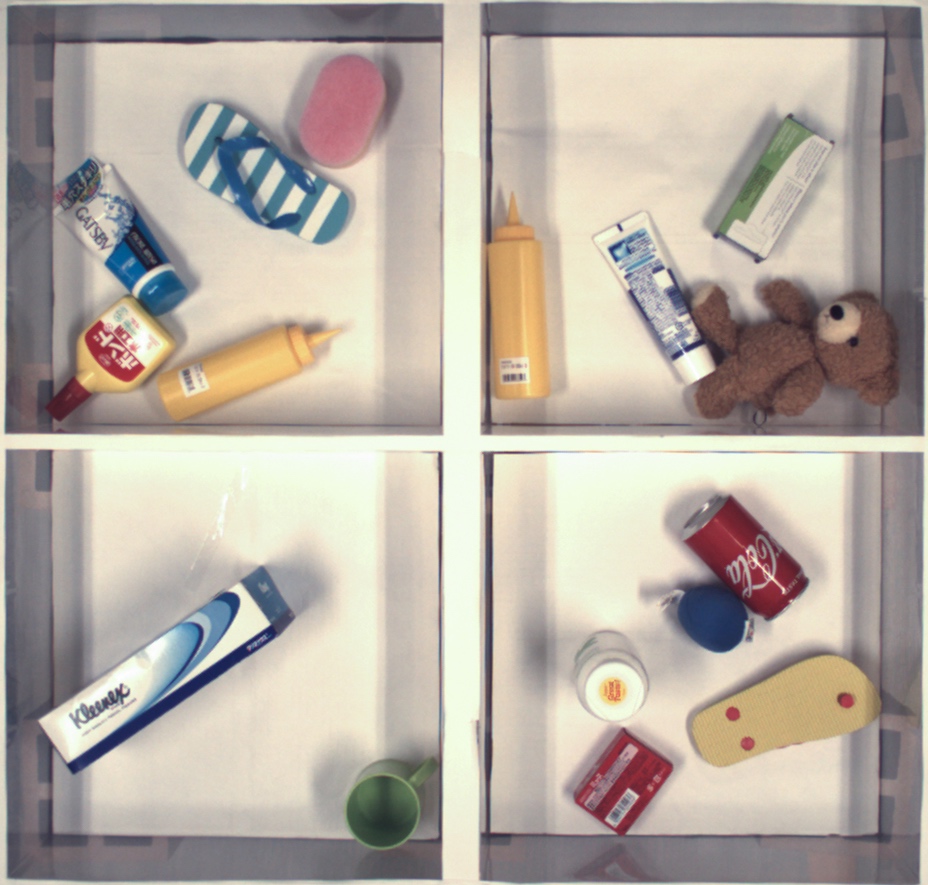}
\end{minipage}
\begin{minipage}{0.47\hsize}
\includegraphics[width=\hsize]{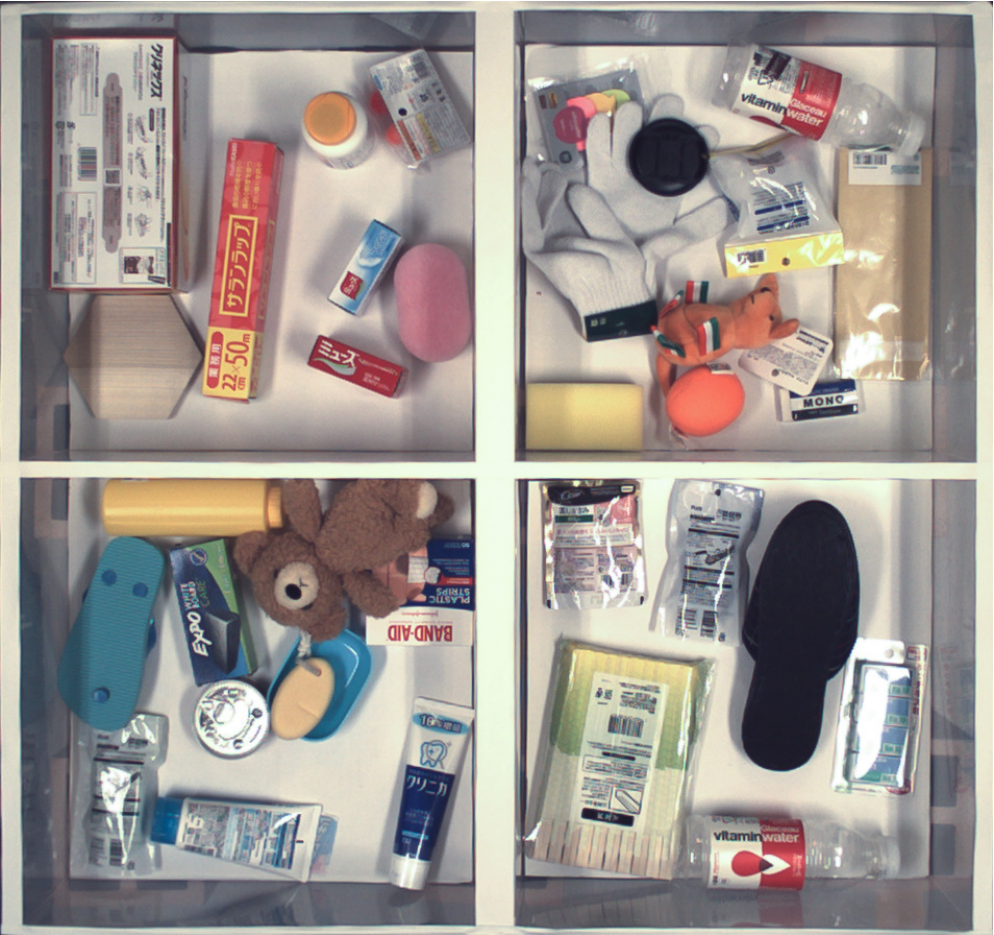}
\end{minipage}
\caption{Examples of average (left) and highly-cluttered (right) environments. The dimension of each box is 400mm $\times$ 405mm.}
\label{fig:environment}
\end{figure}

\begin{figure}
\vspace{2mm}
\begin{center}
\includegraphics[width=0.5\columnwidth]{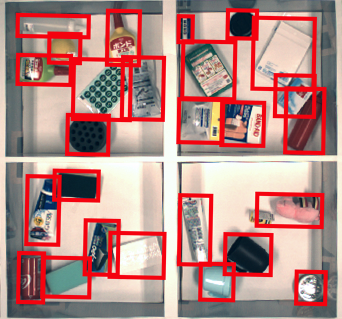}
\caption{Predicted bounding boxes of each object.}
\label{fig:pick-point}
\end{center}
\end{figure}

We created a new dataset, {\it PFN Picking Instructions for Commodities Dataset (PFN-PIC)}, consisting of 1,180 images with bounding boxes and text instructions annotated.
Each image contains 20 objects on average, which are scattered across the four boxes, as in \Cref{fig:environment}.
We chose commodities and daily household items, including many objects which might not look familiar for most people or which many people do not know the specific names of, so that they would have to use indirect expressions referring to their positions and attributes such as color or texture.
Also, we intentionally included multiple objects of the same kind, so that you cannot simply use the name of the object to point to, without referring to the absolute or relative position of the object.
Containing many commodities, we believe our dataset will be useful to train a practical system that is to be used in real-world scenarios.

Each object in the imagery is annotated with a bounding box as shown in Figure~\ref{fig:pick-point}.
We annotated each object in the imagery with text instructions by at least three annotators via crowd sourcing, Amazon Mechanical Turk.
The annotators are asked to come up with a natural language instruction with which the operator can uniquely identify the target object without interpretation ambiguity.
They are also asked to use intuitive, colloquial expressions that they would use when talking to friends.
However, we did not carefully check if every instruction has interpretation ambiguity, deliberately leaving some interpretation ambiguities since one of the main focuses of our work is the interactive clarification process.

In total, we collected 1,180 images with 25,900 objects and 91,590 text instructions.\footnote{The dataset is available at \url{https://github.com/pfnet-research/picking-instruction} (along with a Japanese dataset).}
We split the dataset into a training set consisting of 1,160 images and 25,517 objects, and a validation set consisting of 20 images and 383 objects. 
Since some of the annotations include misspelling and do not appropriately specify target objects, we manually reviewed all the text instructions in the validation set and removed inappropriate instructions; the total number of the objects and the instructions used for validation are 352 and 898, respectively.  
\subsection{Details of Machine Learning Setup}
In this section, we describe details of our machine learning models.
All hyper-parameters described in this section were found through a trial-and-error process on the validation data.

\subsubsection{Candidate Object Detection Model}
We trained a candidate object detection model, which is described in~\Cref{sec:method}, on our dataset.
As a base CNN network~\cite{liu2016ssd} for feature extraction, we used a VGG16 model which was pretrained on ImageNet dataset~\cite{deng2009imagenet}.
The model is trained for 60,000 iterations by SGD (stochastic gradient descent) with momentum.\footnote{We used the mini-batch size of 32, and the learning rate starting from 0.001, decayed by 0.1 every 20,000 iterations.}
Additionally, we employ data augmentation by randomly flipping the images vertically, since our image dataset was taken from a fixed top-down perspective and is different from the horizontal perspective commonly used with existing datasets such as PASCAL VOC~\cite{everingham10pascalvoc}.

\subsubsection{Target Object and Destination Box Selection Model}

We also trained an object comprehension model, described in ~\Cref{sec:method}, on our dataset.
A CNN in the model uses the final layer (after pooling) of 50-layer ResNet~\cite{he2016resnet} pre-trained on ImageNet dataset~\cite{deng2009imagenet}\footnote{We use Chainer~\cite{tokui2015chainer} to import pre-trained ResNet on \url{https://github.com/KaimingHe/deep-residual-networks}, provided by the authors of the ResNet paper.}
Training is performed by minimizing a max-margin loss with a margin $m = 0.1$ between correct sentence--object pairs and randomly-sampled incorrect pairs.
Optimization is performed for 120,000 iterations with Adam.\footnote{We used the mini-batch size of 128 and the learning rate starting from  0.0005, decayed by 0.9 every 4,000 iterations. Dropout is applied with a ratio of 0.1 at each layer of MLP, CNN's output layer, and word embedding layer.}

We construct word vocabulary from words that appear more than once in the training data.
Words that do not appear in the training data are replaced with a special out-of-vocabulary token ``UNK,'' as a common practice used in NLP.
We also used word dropout (dropping tokens themselves) with a ratio of 0.1.
Stochastically dropping the latter half of a sentence is also applied with a ratio of 0.05 during training of the target object selection model.
The number of hidden units in MLP and LSTM is 512 and the number of layers is 3.
These settings are also used for training a destination prediction model based on LSTM and MLP, as described in~\Cref{sec:method}; however, we changed the loss function to cross entropy over the classes (i.e. which box is correct among four boxes) and the number of the units to 256.

\subsection{Robotic System Setup}

\begin{figure}
\centering
\vspace{2mm}
\includegraphics[width=0.9\columnwidth]{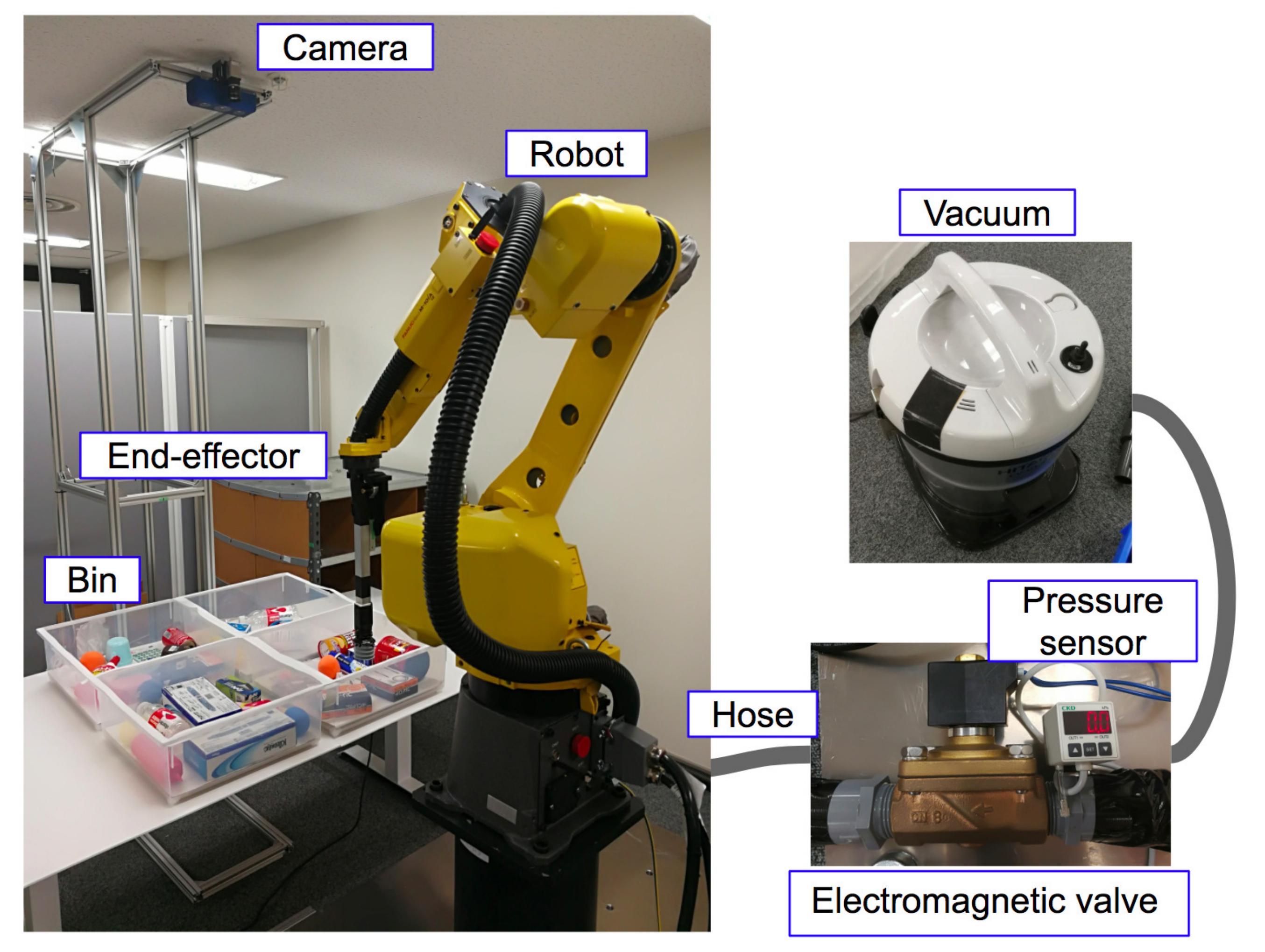}
\caption{Robot setup for experiments}
\label{fig:robotsystem}
\end{figure}

Our robotic system, shown in~\Cref{fig:robotsystem}, consists of an industrial FANUC M10$i$A robot arm equipped with a vacuum gripper as end-effector to handle a variety of objects without taking grasping posture or mechanical stability into consideration.
The vacuum is turned on and off by an electromagnetic valve, through the robot controller.
To validate whether an object has been successfully grasped, we utilize a PPG-CV pressure sensor of CKD corporation.
Furthermore, we use an Ensenso N35 stereo camera in combination with an IDS uEye RGB camera to overlook the workspace of the robot arm and retrieve point clouds of the scene.

To control our robotic system and process data, we use a PC equipped with an NVIDIA GeForce GTX 1070 GPU and an Intel i7 6700K CPU running Ubuntu 16.04 with ROS Kinetic along with our own software modules.
These modules consist of a motion planner, which makes use of a rapidly exploring random tree (RRT) to plan movements of the robot, a manipulation planner capable of querying the inverse kinematics of the robot and construct a path in order to pick and place the target object, interfaces to the cameras, and a task planner that manages all of the other modules.

\section{Results}
\label{sec:results}

\subsection{Software Simulation Results} %

\begin{table}[t]
\begin{center}
\begin{tabular}{ccc}
Candidate object & Destination & Target object \\
detection & box selection & selection \\ \hline
98.6\% & 95.5\% & 88.0\% \\
\end{tabular}
\caption{Performance of each subtask on the validation set. 
The candidate object (bounding box) detection module is evaluated by average precision (AP)~\cite{everingham10pascalvoc}.
The destination box selection is evaluated by top-$1$ accuracy, and the object selection modules is evaluated by top-$k$ accuracies.}
\label{tab:module_performance}
\end{center}
\end{table}

\begin{figure}[t]
\centering
\vspace{2mm}
\includegraphics[width=0.8\columnwidth]{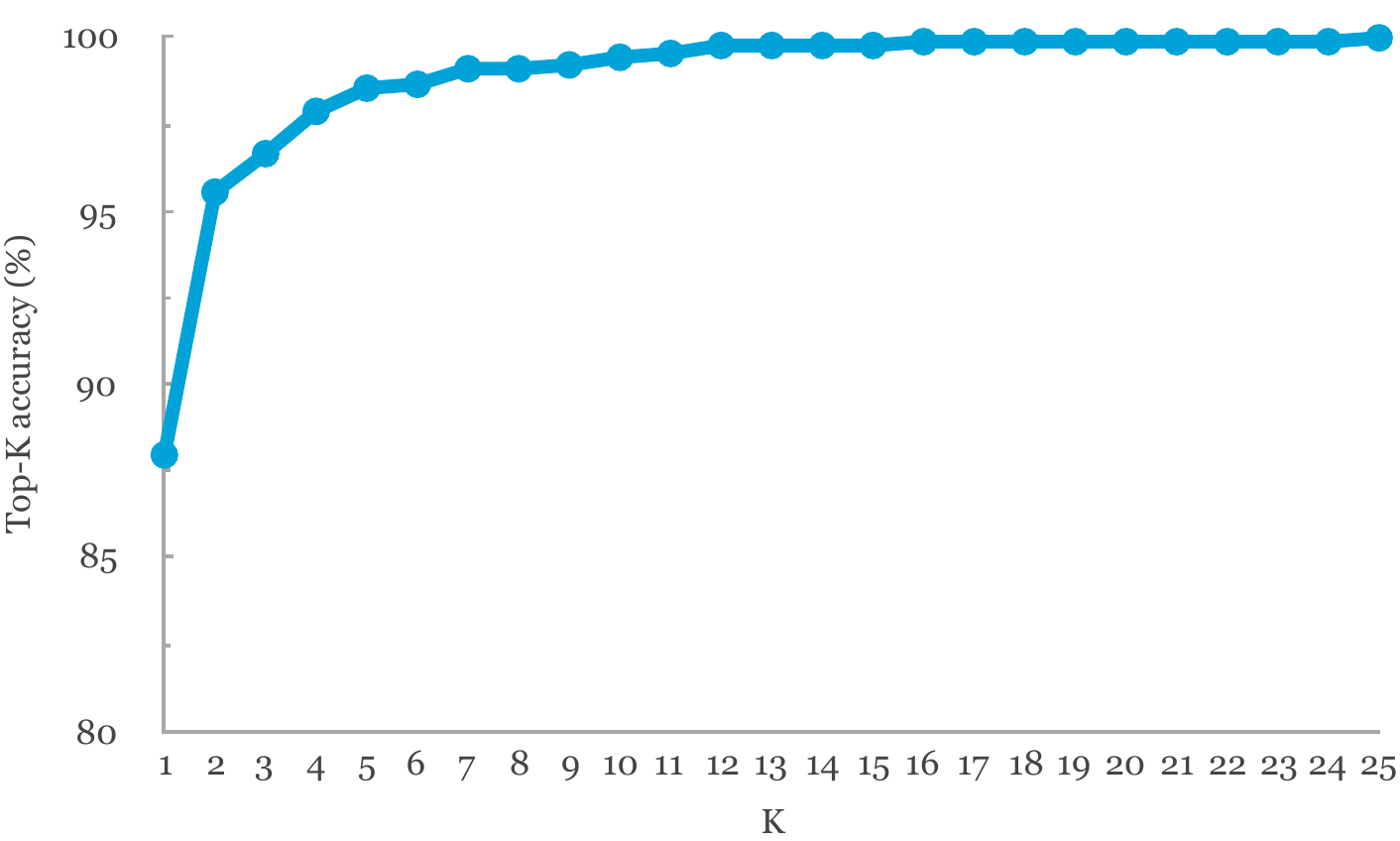}
\caption{Top-$k$ accuracies of target object selection given a spoken instruction.}
\label{fig:topk}
\end{figure}

We first show the performance of each module on the validation data in~\Cref{tab:module_performance}.
Both the candidate object detection and destination box selection modules achieved over 95\% precision and accuracy.
The target object selection module achieved a top-$1$ accuracy of 88\%, which we believe is quite good considering the given instructions are naturally colloquial and an average of 20 objects are placed in each instance.
The accuracies for top-$k$ evaluations are shown in~\Cref{fig:topk}.
When top-$2$ accuracy is considered, the object selection accuracy is largely improved to 95.5\%.
This result suggests that there is a high degree of interpretation ambiguity in our dataset, and getting detailed and elaborated feedback through a clarification process will be important to minimize the detection error in practice.

\begin{table}[t]
\begin{center}
\begin{tabular}{lc}
                              & Target Object Selection \\ \hline
Unambiguous cases only        & 94.9\% \\
Ambiguous cases only          & 63.6\% \\ \hline
Total (without clarification) & 88.0\% \\
Total (with clarification)    & 92.7\% \\ \hline
\end{tabular}
\caption{Comparison of the top-1 target object selection accuracies for unambiguous/ambiguous cases, and the total accuracies with and without the interactive clarification process. The accuracy for ambiguous cases was calculated for the top-ranked object output by the system.}
\label{tab:clarification}
\end{center}
\end{table}

To demonstrate the efficacy of the interactive clarification process, we also conducted a simulated experiment with clarification.
Since our crowd-sourced dataset does not contain clarification sentences even for ambiguous cases, we consider another textual instruction annotated to the same image instance as an additional instruction as a response to our robot's clarification question.\footnote{We pick the instruction that has the least number of word overlap with the original instruction sentence as the clarifying sentence to provide, assuming human operators would avoid re-using the same expressions as in the initial instruction and try to come up with other way of narrowing down the target object and destination box.}
When the given instruction is recognized as \emph{ambiguous} by the robot, we feed the clarification instruction selected this way to the system agent.
It turned out 21\% of the test instances were recognized as \emph{ambiguous} by the system agent.
By providing the additional clarifying instruction for these instances recognized as \emph{ambiguous}, the overall top-$1$ accuracy significantly increased to 92.7\% as shown in \Cref{tab:clarification}, which is 4.7\% higher than the accuracy without the clarification.
This improvement corresponds to an overall error reduction of 39.2\%, showing that providing additional clarifying instructions is highly effective when the instruction given by the human operator has interpretation ambiguity.

We also investigated the relation of the model accuracies and the interpretation ambiguity.
As shown in~\Cref{tab:clarification}, the target object detection selection accuracy was 94.9\% when the instruction was not considered ambiguous, whereas the accuracy for the top-ranked object was 63.6\% when the system judged the instruction as ambiguous.
This result clearly shows that our target object selection model is highly accurate when the given instruction does not have interpretation ambiguity, and a large portion of the errors are for the ambiguous cases.
Therefore, properly resolving the interpretation ambiguity by the iterative clarification process is important to improve the overall accuracy of the target object selection.
We also found that even when the instance was judged as \emph{ambiguous}, our candidate object detection module succeeded in detecting the correct target object as a potential candidate in 90.9\% of the cases.

\subsection{Analysis of Simulation Results}

\begin{figure*}
\vspace{3mm}
\centering
\begin{minipage}[t]{0.24\hsize}
\includegraphics[width=\hsize]{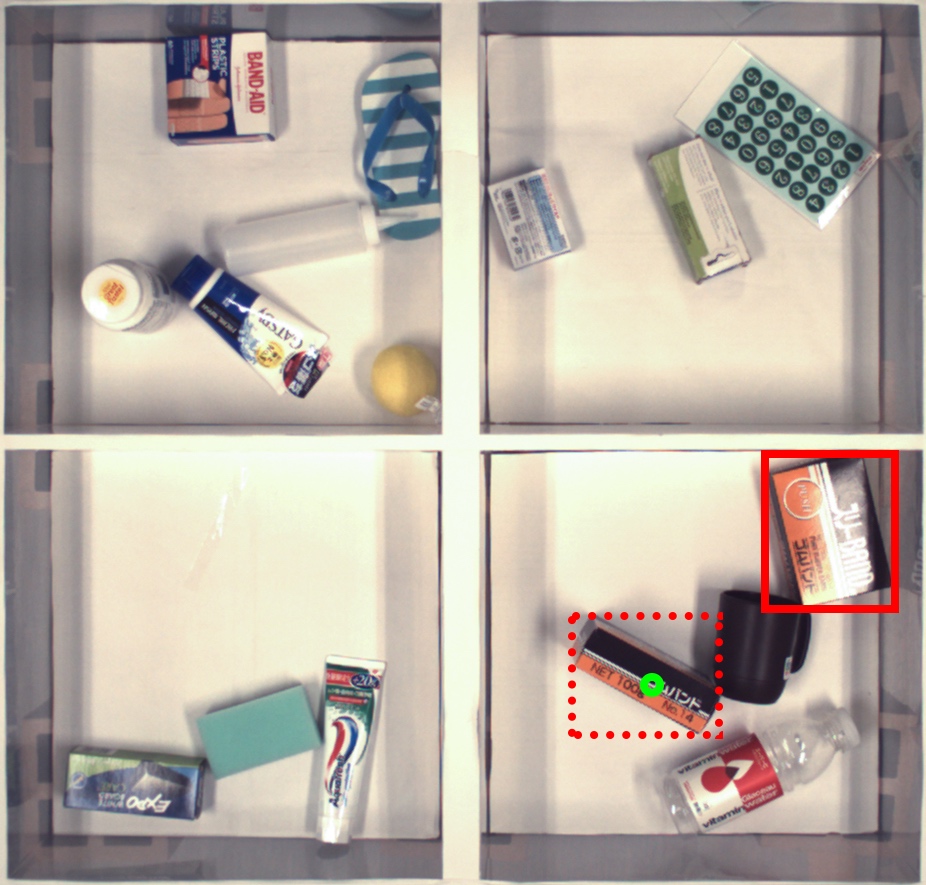} \footnotesize
(a) ``grab the thin orange and black box and put it in the left lower box'' (failure)
\end{minipage}
\begin{minipage}[t]{0.24\hsize}
\includegraphics[width=\hsize]{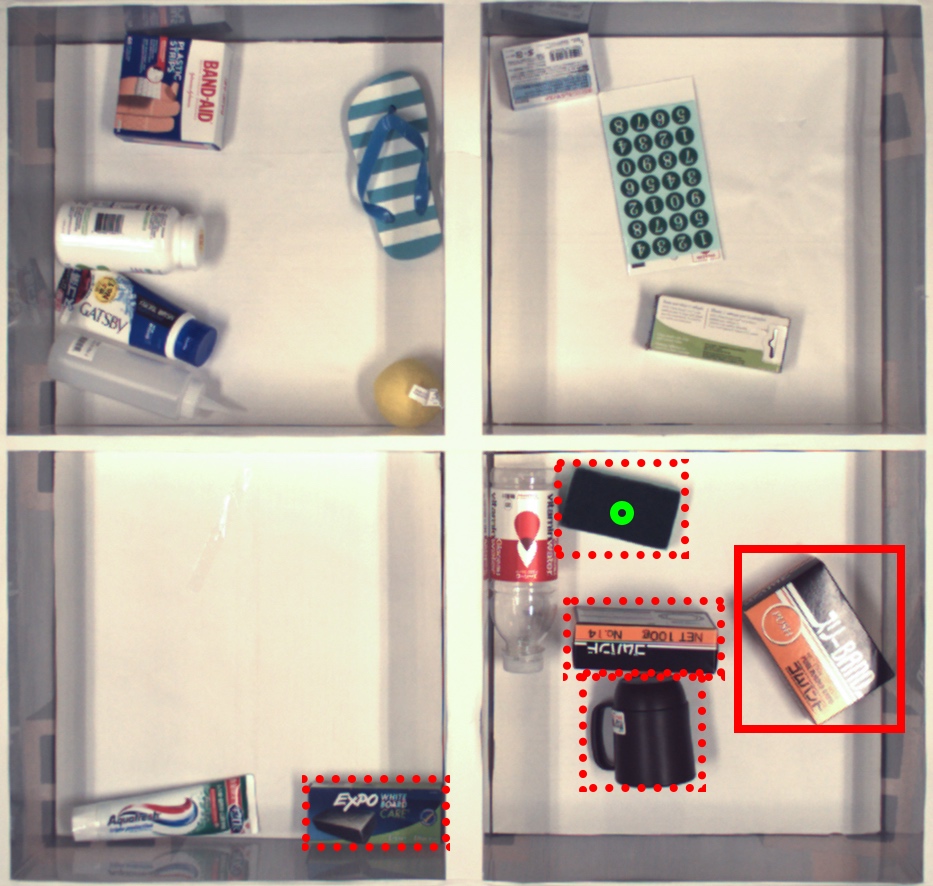} \footnotesize
(b) ``move the lower right side {\it black box} to the upper left hand box'' (failure)
\end{minipage}
\begin{minipage}[t]{0.24\hsize}
\includegraphics[width=\hsize]{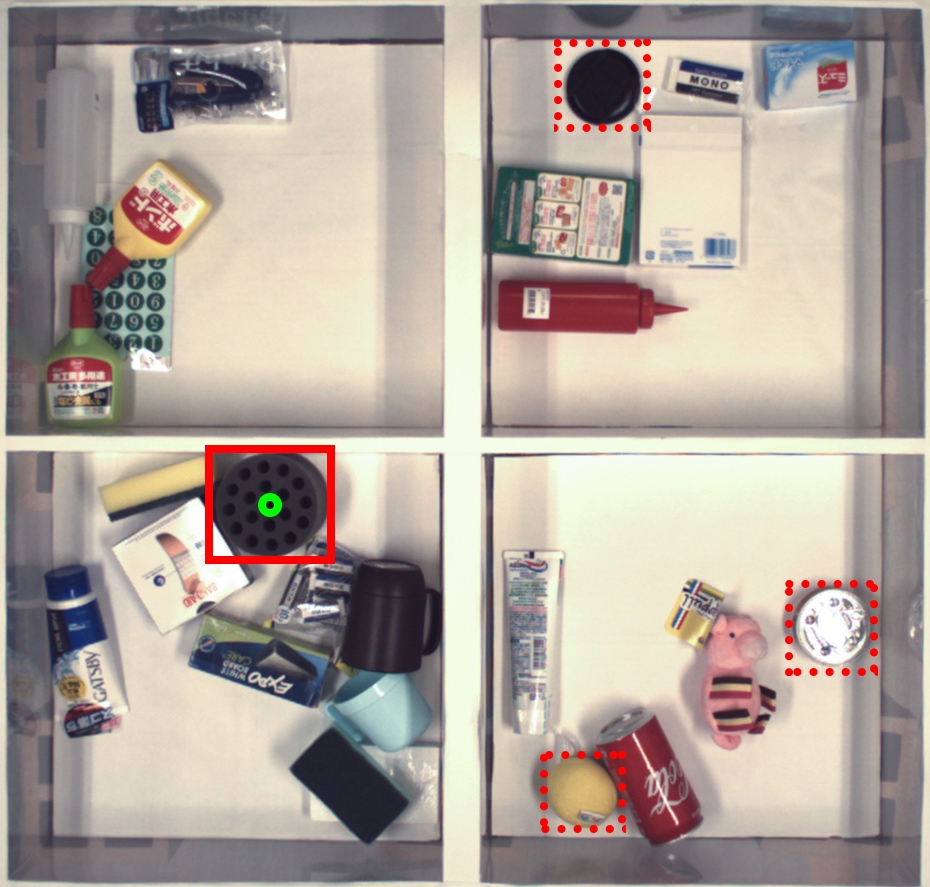} \footnotesize
(c) ``move the round object with multiple holes to upper right box'' (success)
\end{minipage}
\begin{minipage}[t]{0.24\hsize}
\includegraphics[width=\hsize]{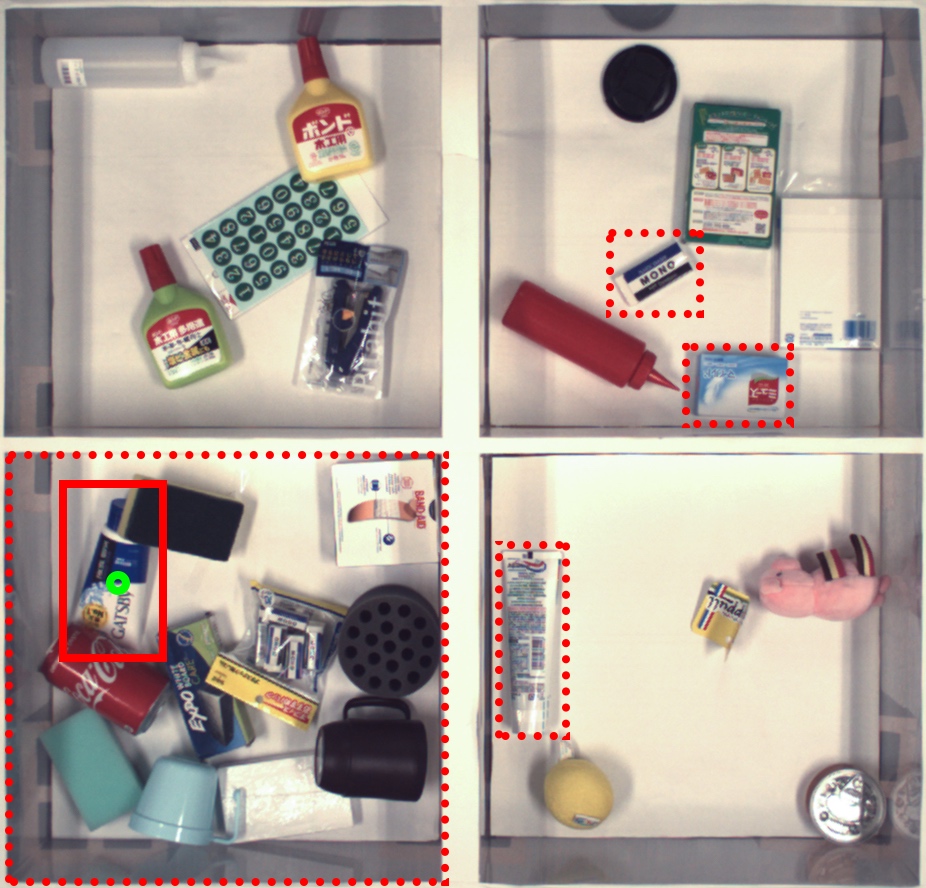} \footnotesize
(d) ``grab the blue and white tube under coke can and move to the right bottom box.'' (success)
\end{minipage}
\caption{Examples of success and failure cases with input images and corresponding text instructions.
The green dot indicates the correct target object, and the red rectangle with a solid line represents the object that the system predicted.
Some regions are also enclosed by a dashed line rectangle to highlight challenges in each instance.
Note that these are not actually predicted bounding boxes.}
\label{fig:example_single}
\end{figure*}

\begin{figure*}
\centering
\begin{minipage}[t]{0.23\hsize}
\includegraphics[width=\hsize]{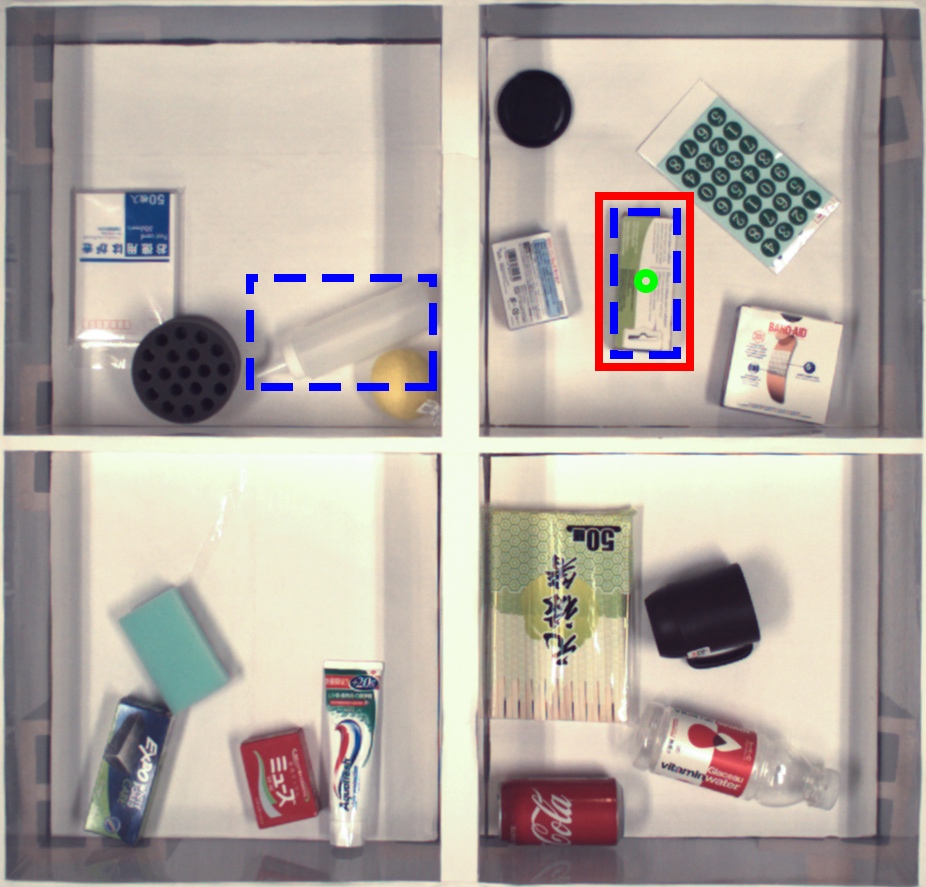}
\end{minipage}
\begin{minipage}[b]{0.23\hsize} \small
1. ``pick the white packet in center and put it into the upper left box'' \vspace{5mm} \\
2. ``move the rectangular object, with a green and white label, located in the middle of the top right box, to the top left box.''
\end{minipage}
\begin{minipage}[t]{0.23\hsize}
\includegraphics[width=\hsize]{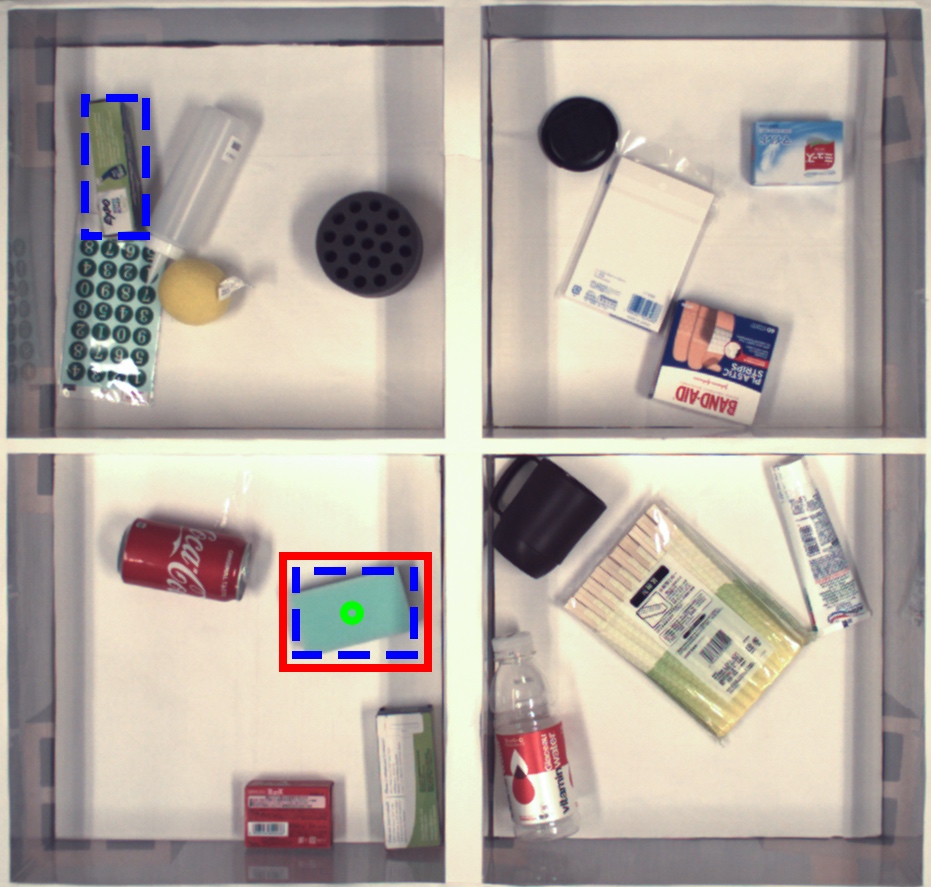}
\end{minipage}
\begin{minipage}[b]{0.23\hsize} \small
1. ``move the blue rectangle the top left box.''  \vspace{5mm} \\
2. ``pick green sponge and put it in the upper box''
\end{minipage}
\caption{Examples of success cases which were judged as ambiguous by the first instructions, but our system could correctly identify the correct object after a clarifying instruction.
Blue rectangles with a dashed line represent ambiguous objects for the first (ambiguous) instruction, and red
rectangle with a solid line represents the final (correct) prediction after clarification.}
\label{fig:example_interactive}
\end{figure*}

In this section, we present some analysis results from our software simulated experiments.
Figure~\ref{fig:example_single} shows four examples of input image and text pairs and detected target objects.
The first two cases, (a) and (b), are failure cases, which are challenging due to the existence of objects with similar appearances.
In case of (a), there are two identical objects (an orange and black box, a box of rubber band) placed differently in the bottom right box.
The correct, upright one was referred to as ``thin,'' but our system failed to understand that refers to the upright box.
The second failure case (b) is related to color.
The operator said ``black box,'' but our system selected a ``black and orange'' box.
This example suggests that if there is an object that is mostly covered by a single color, human operators often ignore other objects that contain the color partially.
For the case of (c), the system correctly recognized the referring expression, ``object with multiple holes,'' and succeeded in identifying the correct object out of multiple similarly shaped objects.
Finally, (d) is a success case where the system correctly recognized the target object even in a highly cluttered environment.
As seen in these examples, our system has succeeded in recognizing the correct target object even for challenging input, but is often confused by similarly shaped or colored objects.

Figure~\ref{fig:example_interactive} shows two cases where the system successfully selected the target object after a clarifying instruction.
In the first case, although the system could not identify the target object because of color ambiguity, (``white packet in center''), the given clarifying instruction properly disambiguated it by being more specific about the color together with its location (``green and white label, located in the middle of top right box'').
In the second case, although the system was unsure about the target object with ``blue rectangle'' in the first instruction, it raised the confidence after receiving the clarifying instruction with ``green sponge''.

\subsection{Physical Robot Experiment Results}

To evaluate the performance of our system in a real-world setting, we also conducted experiments with a physical robot arm and with seven external human operators.\footnote{The human operators consist of three native speakers of American English, one speaker of British English, and three non-native, but fluent speakers of American English.}

We first set up an environment as in the dataset we created (e.g. \Cref{fig:environment}), and ask one of the operators to instruct the robot to move an object to another box.
Our robot can ask the operator to provide additional information when it cannot confidently identify the target object or destination box.\footnote{The feedback can be made up to twice; if the robot still cannot narrow down the target after the feedback is made twice, that trial is regarded as failure, and we move on to the next trial.}
After 5--7 objects have been moved, we refresh the environment by replacing all objects with another set of randomly selected objects.
We performed a total of 97 object picking trials, out of which 63 trials were performed with only known objects, and 34 trials were performed by mixing known objects and approximately 30\% of unknown objects that are not in our training data.

\begin{table*}
\vspace{3mm}
\begin{center}
\begin{tabular}{rccccc}
                        & Destination    & Target Object  & Pick and Place & Pick and Place & Avg. Number  \\
                        & Box Selection  & Selection      & (only)         & (end-to-end)   & of Feedback  \\ \hline
Without unknown objects & 88.9\% (56/63) & 77.8\% (49/63) & 98.0\% (48/49) & 76.2\% (48/63) & 0.41 (26/63) \\
   With unknown objects & 91.2\% (31/34) & 70.6\% (24/34) & 95.8\% (23/24) & 67.6\% (23/34) & 0.53 (18/34) \\ \hline
                  Total & 89.7\% (87/97) & 75.3\% (73/97) & 97.3\% (71/73) & 73.1\% (71/97) & 0.45 (44/97) \\
\end{tabular}
\caption{Experimental results with a physical robot arm.
\textup{\emph{Destination Selection} and \emph{Target Object Selection} correspond to our destination box and target object selection accuracies.
\emph{Pick and Place (only)} and \emph{Pick and Place (end-to-end)} respectively correspond to the success rate of our object picking and placing task calculated only for successfully-detected instances (\emph{only}) and that for all instances (\emph{end-to-end}), including those in which the target box or object detection has failed.
\emph{Avg. Number of Feedback} indicates the average number of per-session clarification questions asked by the robot.}}
\label{tab:robot-experiment}
\end{center}
\end{table*}

\Cref{tab:robot-experiment} shows the success rates of our picking task as well as the accuracies for the destination box selection, target object selection, and the average number of per-session clarification questions asked by the robot.
Even with our challenging setting with external human operators who do not have any prior knowledge about our task, our target object selection module achieved a 75.3\% top-1 accuracy, and the end-to-end success rate of the picking and placing task was 73.1\%.
Even with a highly challenging setting with 30\% of unknown objects mixed, our recognition model still works reasonably well, with a 70.6\% object selection accuracy, and the end-to-end success rate of picking and placing was 67.6\%.
One reason for such generalization power is that our model used CNNs which are pretrained on ImageNet.
Please note we told the human operators to avoid using command-like instructions, and to use colloquial expressions they would use when talking to a friend; we did not ask them to use specific patterns or expressions.

The target object selection accuracy of 75.3\% was lower than the software simulated accuracy of 88.0\%.
We found this is primarily because the environment used in our robot experiments is slightly different from the original environment where the dataset was taken.
Especially, the differences in the camera position, color temperature, and lighting conditions had a large impact on the accuracy.
Especially, the color temperature had the largest impact, and instructions that include color information as perceived by the instructor (e.g., orange) did not always match the color as perceived by our system or the annotated color in the dataset (e.g., red).
We will try to fill in this gap and improve the accuracy as future work.

Although our picking and placing accuracy alone was as high as 97.3\%, we found the grasping ability of the vacuum end-effector was sometimes problematic.
The versatility of a vacuum end-effector allows the robot to grasp various kinds of objects without precise grasping points, but some of the objects were still difficult to grasp.
For example, sponges could not be sucked reliably, and the strong suction power for the vacuum sometimes caused empty beverage cans to dent, classified as a failure case of grasping.

\section{Conclusion}
\label{sec:conclusion}

In this paper, we proposed the first robotic system that can handle unconstrained spoken language instructions and can clarify a human operator's intention through interactive dialogue.
Our system integrated state-of-the-art technologies for object recognition and referring expression comprehension models, and we also proposed to make several modifications so that the system can handle previously unseen objects effectively.
We evaluated our system in a highly challenging, realistic environment, where miscellaneous objects are randomly scattered, and we not only achieved a high end-to-end picking accuracy of 73.1\% with an industrial robot, but also demonstrated that the interactive clarification process is effective for disambiguation of a human operator's intention, reducing the target object selection error by 39\%.
We believe our system's performance for colloquial spoken instructions and cluttered environments as we focused on in this work will serve as a benchmark for future development of robotic systems that interact with humans via spoken languages.

For future work, we plan to extend our method to support multiple languages using the same architecture, with an attempt to share knowledges across different languages to increase the accuracy and coverage, since our system is designed in a totally language agnostic way.
Moreover, we also plan to improve our picking accuracy by designing a grasping strategy with a combination of a vacuum and gripper for the end-effector.

\section*{ACKNOWLEDGMENT} \small
We would like to thank Masaaki Fukuda, Totaro Nakashima, Masanobu Tsukada, and Eiichi Matsumoto for their assistance with our extensive data collection.
We are also grateful to anonymous reviewers and Jason Naradowsky for insightful comments.

\bibliographystyle{IEEEtran}
\bibliography{IEEEabrv,bibliography}

\end{document}